\documentclass{article}

\usepackage{arxiv}

\usepackage[utf8]{inputenc} 
\usepackage[T1]{fontenc}    
\usepackage{hyperref}       
\usepackage{url}            
\usepackage{booktabs}       
\usepackage{amsfonts}       
\usepackage{nicefrac}       
\usepackage{microtype}      
\usepackage{graphicx}
\usepackage{natbib}
\usepackage{doi}
\usepackage{multirow}
\usepackage{amsmath}
\usepackage{array} 
\newcolumntype{C}[1]{>{\centering\arraybackslash}p{#1}} 

\title{Real-Time Bus Departure Prediction Using Neural Networks for Smart IoT Public Bus Transit}

\author{
	{\hspace{1mm}Narges Rashvand} \\
	Department of Electrical and Computer Engineering\\
	The University of North Carolina at Charlotte\\
	Charlotte, NC, USA \\
	\texttt{nrashvan@uncc.edu} \\
 \And
	{\hspace{1mm}Sanaz Sadat Hosseini } \\
	Department of Civil and Environmental Engineering\\
	University of North Carolina at Charlotte\\
    Charlotte, NC, USA \\
	\texttt{shossei7@uncc.edu} \\
  \And
	{\hspace{1mm}Mona Azarbayjani} \\
	School of Architecture\\
	The University of North Carolina at Charlotte\\
	Charlotte, NC, USA \\
	\texttt{mazarbay@uncc.edu} \\
  \And
	{\hspace{1mm}Hamed Tabkhi} \\
	Department of Electrical and Computer Engineering\\
	The University of North Carolina at Charlotte\\
	Charlotte, NC, USA \\
	\texttt{htabkhiv@uncc.edu} \\
}

\begin{document}
\maketitle

\begin{abstract}
Bus transit plays a vital role in urban public transportation but often struggles to provide accurate and reliable departure times. This leads to delays, passenger dissatisfaction, and decreased ridership, particularly in transit-dependent areas. A major challenge lies in the discrepancy between actual and scheduled bus departure times, which disrupts timetables and impacts overall operational efficiency.
To address these challenges, this paper presents a neural network-based approach for real-time bus departure time prediction tailored for smart IoT public transit applications. We leverage AI-driven models to enhance the accuracy of bus schedules by preprocessing data, engineering relevant features, and implementing a fully connected neural network that utilizes historical departure data to predict departure times at subsequent stops. In our case study analyzing bus data from Boston, we observed an average deviation of nearly 4 minutes from scheduled times. However, our model, evaluated across 151 bus routes, demonstrates a significant improvement, predicting departure time deviations with an accuracy of under 80 seconds. This advancement not only improves the reliability of bus transit schedules but also plays a crucial role in enabling smart bus systems and IoT applications within public transit networks. By providing more accurate real-time predictions, our approach can facilitate the integration of IoT devices, such as smart bus stops and passenger information systems, that rely on precise data for optimal performance.

\end{abstract}

\keywords{feature engineering \and deep learning \and  fully connected neural networks \and  bus departure time prediction  \and IoT }

\section{Introduction}
\label{sec:intro}  
Public bus transit in the United States has faced significant challenges over recent years, marked by a decline in ridership due to various factors. These include unreliable bus services, inadequate bus stop coverage, safety concerns,
the broader availability of affordable personal vehicles, ride-hailing options, and low gas prices \cite{brough2021understanding, covington2018overcoming, lee2022whats, erhardt2022why, alltransit, shi2021impact, diab2015bus, streetsblog2020buses, tsg2021transport}.

These factors have led to increased traffic congestion, higher carbon emissions, and negative environmental impacts. The COVID-19 pandemic exacerbated the situation, leading to a sharp drop in transit usage, with bus ridership in many cities falling to just 62\% of pre-pandemic levels by mid-2022 \cite{ziedan2023transit, apta2024ridership, brough2021understanding, qi2023impacts, bi2024can, tirachini2020covid, rashvand2023real, jiao2013transit, oshyani2014realtime, hosseini2024demographic}.

\begin{figure}[h]
\centering
\vspace{0pt}
\includegraphics[width=4.8in]{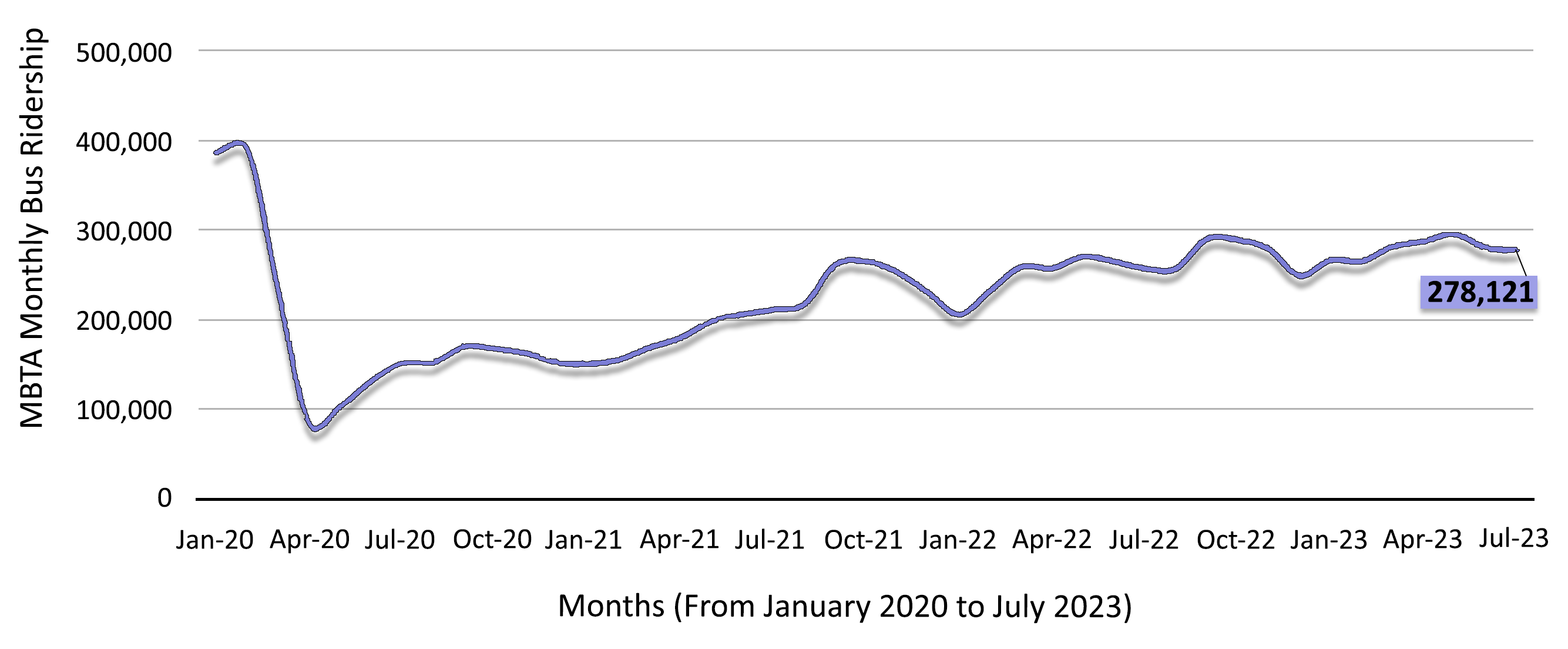}
\caption{MBTA monthly bus ridership since January 2020, showing the current decline of 61.75\% in ridership, compared to pre-pandemic levels \cite{avisonyoung2024}.}
\label{boston_bus_ridership}
\end{figure}

\textls[-15]{Bus ridership in Boston, managed by the Massachusetts Bay Transportation Authority (MBTA), reflects these broader national trends. In 2019, MBTA bus ridership averaged approximately
727,382 unlinked passenger trips per month. However, as shown in Figure \ref{boston_bus_ridership}, the pandemic caused a significant decline in ridership, with average monthly trips dropping to 278,121 in July 2023, approximately 61.75\% lower than pre-pandemic levels. Although there has been a gradual recovery, the MBTA has lagged behind other major transit agencies in the U.S. in regaining ridership \cite{axios2023, apta2024ridership, nbc2024, avisonyoung2024, pioneerinstitute2024}.}

The U.S. public bus transit system has untapped potential to serve urban commuters effectively, being more reliable, cost-effective, and flexible.
Nearly every major city in the United States offers some form of bus service, with many operating 24 hours a day. Addressing the challenges mentioned earlier requires innovative solutions that integrate technology, reflecting the resilience of public transit systems and the importance of continuous investment and innovation in this sector \cite{hosseini2023towards, rashvand2023real, hosseini2024demographic}.

Despite the challenges discussed, the majority of bus services struggle with reliability, particularly in delivering accurate arrival/departure times. The disparity between scheduled and actual bus arrival/departure times has consistently been identified as a major reason why commuters avoid using bus transit systems in many cities \cite{basak2019data}.  
Unreliable transit services can disrupt commuters' plans, leading to missed rides and longer waits,  consequently resulting in decreased bus ridership and greater use of alternative transportation methods \cite{rashvand2023real, basak2019data}. To tackle schedule unreliability and improve on-time performance, real-time bus arrival/departure prediction supported by IoT sensors and advanced data analytics is crucial. These technologies, including weather monitors, GPS trackers, and speed detectors, help cities enhance transit scheduling and improve reliability, efficiency, and passenger satisfaction \cite{hosseini2024demographic, rashvand2023real, basak2019data, jabamony2020iot, abdi2021review}.

\begin{figure}[h]
\centering
\includegraphics[trim= 0pt 0pt 0pt 10pt , clip, width=4.3in]
{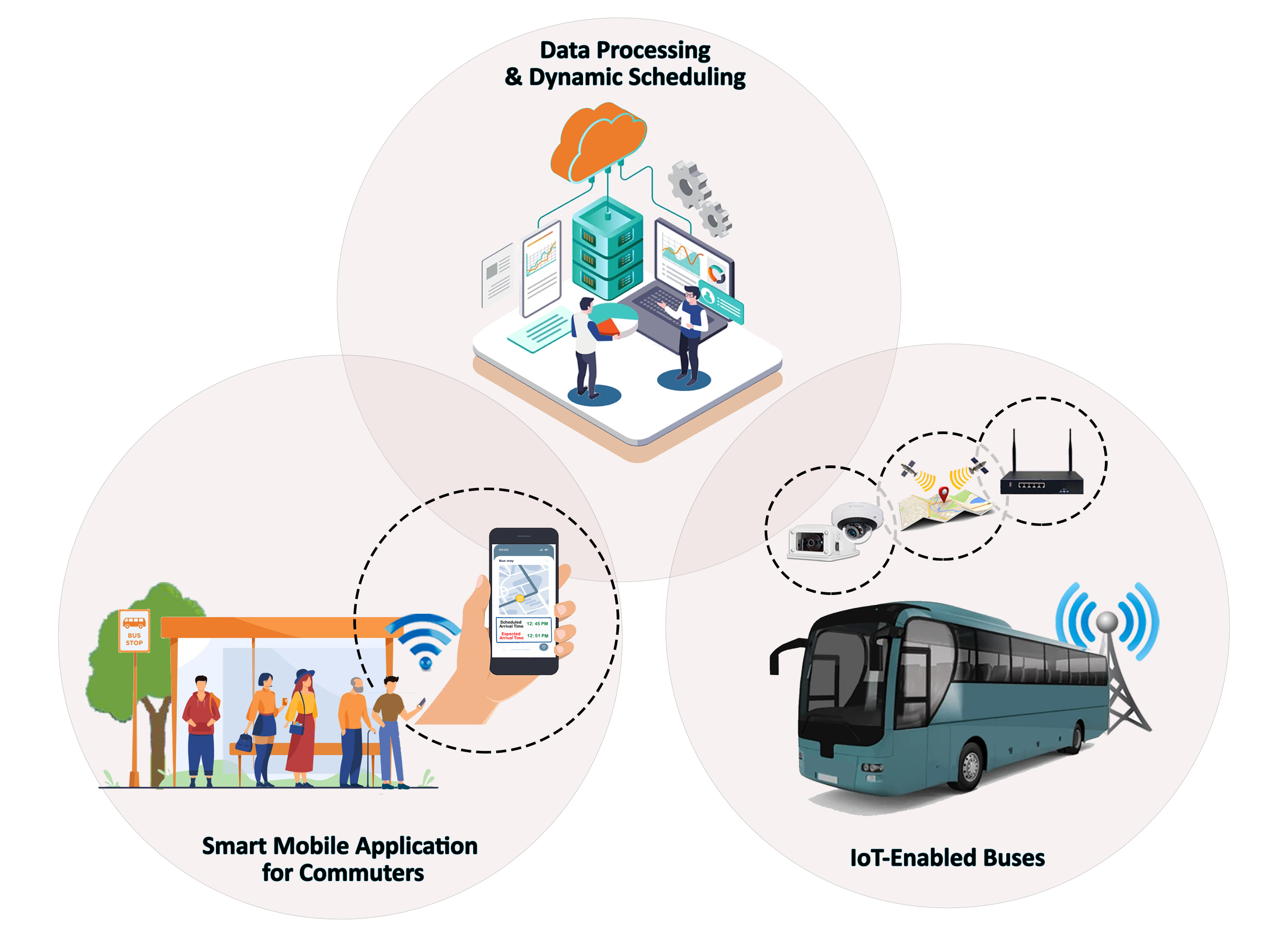}
\caption{Comprehensive IoT-enabled smart bus system: This diagram illustrates a public transit enhancement framework with three main components: data processing and dynamic scheduling (managed by a centralized cloud platform that processes real-time transit data and employs an AI-powered prediction model to adjust bus schedules dynamically), IoT-enabled buses (which continuously exchange data with the cloud to provide up-to-date information and improved service quality), and a mobile app for passengers to access the bus departure time information.} 
\label{Iot}
\end{figure}

By leveraging advanced technology and AI-driven models, this study aims to enhance the public bus transit system, making it more responsive, efficient, and reliable. Integrating bus departure time prediction models into the existing transportation framework can significantly improve overall service quality by reducing passenger wait times and providing more dependable scheduling for bus departures. Presented in Figure \ref{Iot}, the proposed smart bus system enhances existing bus service infrastructure with its intelligent features. In this ecosystem, each bus functions as an IoT device, continuously exchanging data with the cloud, which hosts an artificial intelligence (AI)-powered bus departure time prediction model. Passengers can access real-time bus departure time information via a mobile app that captures current transit demands. A centralized cloud platform dynamically adjusts schedules based on various features detailed in subsequent sections. Key elements of this framework include seamless integration with current infrastructure, a commuter-centric app, and a cloud-based platform \cite{hosseini2023towards,rashvand2023real}.

\begin{figure}[h]
\centering
\vspace{0pt}
\includegraphics[width=\linewidth]{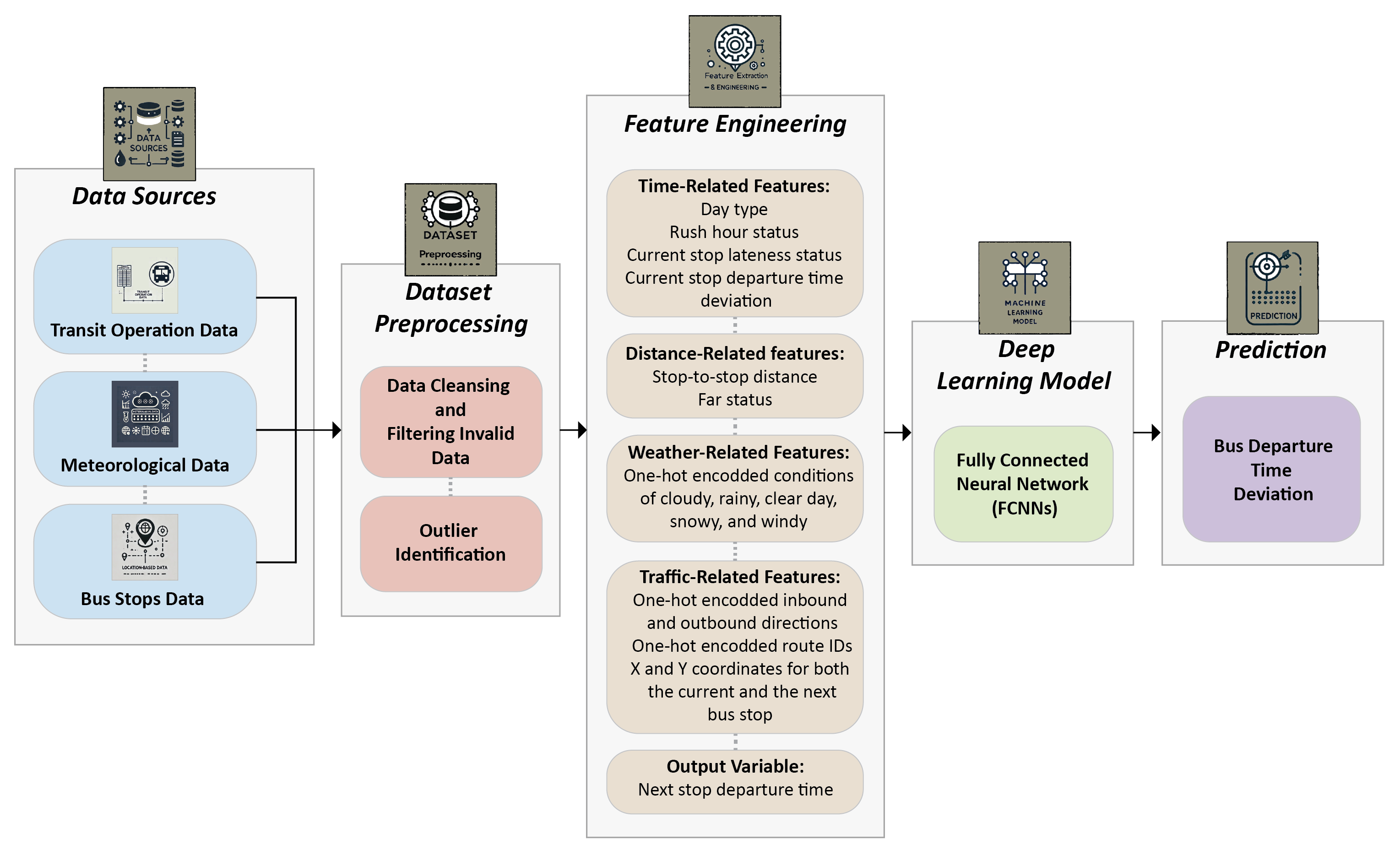}
\caption{An overview of the bus departure time prediction system: The system integrates three key data sources (transit operations data, meteorological data, and bus stop data). The preprocessing stage involves removing invalid entries and detecting outliers. Feature extraction generates both input and output features for the model. An FCNN with a one-bus stop lookback window is trained using the MSE loss function. In the prediction stage, the system provides real-time estimates of bus departure time deviations.} 
\label{FlowChart}
\end{figure}

Therefore, the primary focus of this study is to utilize real-time data for creating a bus departure time prediction model based on deep learning approaches. This platform is designed to provide timely information on bus departure times to passengers via mobile apps and passenger information services, as well as to transit controllers to implement proactive operational strategies. In summary, the contributions of this paper can be outlined as follows:

\begin{itemize}
\item The integration of multiple datasets, transit operations data, meteorological data, and bus stop data,
to enhance predictive accuracy. By combining these three datasets, the present study leverages diverse sources of information to improve the predictions. Comprehensive preprocessing was performed on these datasets to ensure data consistency. Additionally, feature extraction techniques were applied to the integrated data, allowing for the generation of relevant input features that significantly enhance model performance, as illustrated in Figure \ref{FlowChart}.
\item The introduction of a unified fully connected neural network (FCNN) with a one-bus stop lookback window aimed at predicting bus departure times across numerous bus lines within a single bus transit network. This approach effectively utilizes the FCNN's capability to capture temporal dependencies by processing relevant input features.
\item Optimizing the model's size and performance for deployment on IoT devices, balancing accuracy with resource efficiency. 
\end{itemize}

The rest of this paper is structured as follows: Section \ref{sec2} reviews some of the related works in the field, and Section \ref{sec3} details the datasets used. Section \ref{sec4} explains the methodology, covering data preprocessing, feature extraction, and model implementation. Section \ref{sec5} presents the experimental results, and finally, Section \ref{sec6} concludes this paper with the key findings and future research directions.

\section{Related Works} \label{sec2}
Basak et al. (2019) \cite{basak2019data} emphasize that ensemble learning methods, which combine multiple prediction algorithms, can significantly improve prediction accuracy. Their study introduces the Boruta feature selection algorithm to identify the most important input features and compares the performance of different ensemble learning methods, including bagging, boosting, and stacking. They present a data-driven approach to optimize public transit schedules, focusing on maximizing the probability of bus arrivals at timepoints within a desired on-time range, thereby improving overall on-time performance \cite{basak2019data}. Several studies, including Rashvand et al. (2023) and Yin et al. (2017), have developed deep learning models for predicting bus arrival and travel times \cite{rashvand2023real, yin2017prediction, 9529328}. Rashvand et al. (2023) propose a deep learning approach for real-time bus arrival prediction, utilizing a fully connected neural network (FCNN) model evaluated on New York bus transportation data. Their study demonstrates superior scalability and generalization capabilities compared to traditional machine learning models like support vector regression \cite{rashvand2023real}. Moreover, Yin et al. (2017) present a new prediction model using support vector machines (SVMs) and artificial neural networks (ANNs) to predict bus arrival times at stops with multiple routes. The study, conducted using data from Zigong, China, shows that both SVM and ANN models have high accuracy, with the ANN model performing better. Their study highlights the importance of using multiple routes passing the same stop as inputs to improve prediction accuracy \cite{yin2017prediction}.


{Serin et al. (2022) \cite{serin2022predicting} propose a three-layer architecture for predicting bus travel times using machine learning methods. The first layer processes raw data to generate an initial prediction, the second layer predicts residuals between the actual and predicted values, and the third layer integrates these results to produce the final prediction. The models's performance was evaluated using public transportation data from Istanbul, Turkey \cite{serin2022predicting}. Moreover, Jabamony and Shanmugavel (2020) \cite{jabamony2020iot} present an IoT-based smart public transport system (SPTS) that utilizes artificial neural networks (ANNs) to predict bus arrival times. Their system gathers real-time data from IoT-enabled devices such as GPS, traffic, and weather conditions, and trains the ANN model to capture complex patterns affecting bus arrivals. The results show a significant improvement in prediction accuracy, proving the value of combining IoT and ANNs for public transit systems \cite{jabamony2020iot}. Furthermore, Fadaei Oshyani and Cats (2014) \cite{oshyani2014realtime} introduce a hybrid prediction scheme that integrates schedule-based, instantaneous, and historical data to predict real-time bus departure times. This scheme was tested on three trunk bus lines in Stockholm, Sweden, and demonstrated a significant reduction in mean absolute error (MAE) for both operators and passengers when compared to traditional methods, showing the potential of hybrid models for improving bus arrival and departure predictions \cite{oshyani2014realtime}.}

Effective feature selection plays a crucial role in optimizing the performance of deep learning models \cite{zibaeirad2024comprehensive, rashvand2024enhancing}. Bai et al. (2015) propose a dynamic travel time prediction model using SVMs and a Kalman filtering-based algorithm to dynamically adjust predictions with the latest bus operation information, further demonstrating the importance of feature engineering in improving prediction accuracy \cite{bai2015dynamic}.

\section{Materials} \label{sec3}
{This section outlines the key attributes of the three distinct datasets used to enhance prediction accuracy, including transit operations data, meteorological data, and bus stop data.}

\subsection{Transit Operations Data}
For this study, transit operations data were collected from the MBTA Bus Departure Times 2023 dataset \cite{MBTA2023}, which records the departure events for buses in Boston throughout the year 2023. This dataset is publicly available in CSV format from the MBTA and encompasses data from various bus routes.

Each data point in the dataset includes information formatted into 13 fields, including Service Date, Route ID, Direction ID, Half-Trip ID, Stop ID, Timepoint ID, Timepoint Order, Point Type, Standard Type, Scheduled Time, Actual Time, Scheduled Headway, and Headway, 
as described below:
\label{sec:data}  
\begin{itemize}
\item Service Date refers to the date when the trip occurred.
\item Route ID serves as the unique identifier for each route.
\item Direction ID indicates whether the trip is inbound or outbound.
\item Half-Trip ID uniquely identifies the one-way trip.
\item Stop ID identifies the specific stop in the General Transit Feed Specification (GTFS) format.
\item Timepoint ID is the code assigned to a specific bus stop within a particular trip. 
\item Timepoint Order indicates the sequence of this timepoint in the trip.
\item Point Type specifies whether the stop is the starting point, midpoint, or the endpoint for the trip.
\item Standard Type identifies whether the trip should be evaluated on the schedule standard or headway standard. 
\item Scheduled Time denotes the scheduled departure time of the trip.
\item Actual Time indicates the time when the trip actually departed from the timepoint.
\item Scheduled Headway represents the scheduled time gap between this trip and the previous trip at the stop.
\item Headway represents the actual time gap between this trip and the previous trip at the stop.
\end{itemize}

This dataset was selected for its robust characteristics, including a large number of data points, which is ideal for training deep learning models that require extensive data, as well as comprehensive records of actual and scheduled departure times for numerous bus routes. Therefore, we concentrate on the first three months of 2023. This period provides a sufficient volume of data for model evaluation, balancing both training capabilities and resource limitations, while also capturing a diverse range of scenarios within those months. Looking ahead, since the dataset spans the entire year, future research could explore predicting departure times by incorporating seasonal features and variations. 
The characteristics of the original dataset before preprocessing are detailed in Table \ref{tab:dataset_characteristics}. It includes approximately 6 million data points, tracking a total of 940,908 trips. The dataset covers 151 routes and 1111 individual bus stops.

\begin{table}[h]
\centering
\caption{Dataset 
characteristics: insights into MBTA Bus Departure Times 2023---original data and preprocessing overview.}
\label{tab:dataset_characteristics}
\resizebox{\columnwidth}{!}{%
\begin{tabular}{>{\centering\arraybackslash}p{8cm}>{\centering\arraybackslash}p{4cm}>{\centering\arraybackslash}p{4cm}}
\toprule
\textbf{}  
&\textbf{Original MBTA Bus Departure Times 2023} \cite{MBTA2023}
& \textbf{Preprocessed MBTA Bus Departure Times 2023} \\ 
\midrule
\textbf{Number of data points 
}
& 6,361,455 
& 5,954,486 \\
\midrule
\textbf{Number of tracked trips} 
& 940,908 
& 921,829\\
\midrule
\textbf{Number of total bus routes} 
& 151 
& 151 \\
\midrule
\textbf{Number of bus stops} 
& 1111 
& 1111 \\
\midrule
\textbf{Mean of departure time deviation} 
& 261.84 s 
& 211.866 s \\
\midrule
\textbf{Standard deviation of departure time deviation} 
& 309.996 s 
& 189.016 s \\
\bottomrule
\end{tabular}%
}
\end{table}

\subsection{Meteorological Data}
{Understanding weather patterns can be important for accurate model predictions. Adverse weather conditions, such as heavy rain or snow, can impact traffic flow and passenger behavior, leading to delays. To account for these factors, we obtained the meteorological data from the Visual Crossing API \cite {weatherapi}, which offers historical weather information for cities over specified periods. For our analysis, we collected weather data for these three months through this API. These data include details such as temperature, humidity, wind speed, and various weather conditions.}

\subsection{Bus Stop Data}
\begin{figure}[h]
\centering
\vspace{0pt}
\includegraphics[trim=650pt 280pt 2250pt 215pt, clip, width=3.2in]{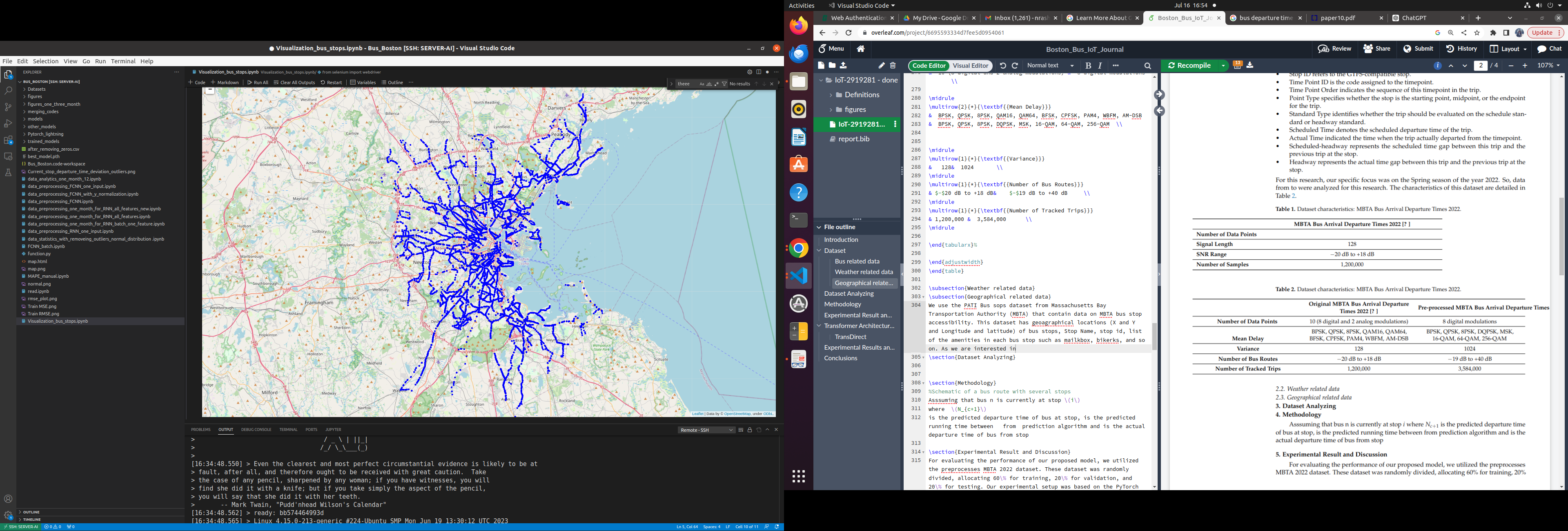}
\caption{\textls[-15]{A visualization of the geographical locations of 1,111 bus stops from the MBTA Bus Departure Times 2023 dataset. The map illustrates the distribution of these bus stops throughout the city of Boston.}}
\label{fig_0}
\end{figure}

{We utilize the MBTA Bus Routes and Stops dataset \cite{stopsdata}, which provides detailed information about MBTA bus stops. Our primary objective is to extract the geographical locations of bus stops from this dataset, as the MBTA Bus Departure Times 2023 dataset does not contain this information. After the preprocessing discussed in the Methodology Section, we use these geographical details as input for our model.
As shown in Figure \ref{fig_0}, this dataset includes the geographical locations of 1111 bus stops from the MBTA Bus Departure Times
2023 dataset and
provides essential details 
such as the geographical coordinates (X and Y) of each bus stop, stop name, and unique stop IDs compatible with GTFS standards.} 
\newline

\vspace{0pt}

\section{Methodology} \label{sec4}
{The purpose of this section is to provide an overview of the key steps of dataset preprocessing and feature engineering, which are crucial for improving the accuracy of our prediction model. Before we explore the details of data cleaning and preprocessing, it is important to highlight some key considerations and the relevance of our work to the field of IoT. Our research investigates the connection between IoT technologies and transportation systems by utilizing data from IoT devices like GPS and sensors on buses. While we do not directly implement any IoT-based technologies, we focus on real-time data collection and employ detailed data cleaning and preprocessing techniques to ensure data quality and integrity, which are crucial for accurate predictions. By integrating diverse features such as time, distance, weather, and traffic, our approach demonstrates how IoT enhances the modeling process through comprehensive data integration. Additionally, we aim to develop a compact, fully connected neural network that balances strong predictive capabilities with computational efficiency, making it suitable for IoT-based transportation applications.
These steps, shown in Figure \ref {FlowChart}, include cleaning the data, detecting outliers, and creating features that capture important patterns in the data. The following subsections provide a detailed overview of these processes.}

\subsection{Dataset Analyzing and Preprocessing}
The analysis starts with initial data cleaning and preprocessing so we can conduct statistical analysis and extract meaningful insights from the dataset. While the original dataset consists of 6,868,730 data points, not all of them are considered valid observations. After filtering out invalid values, the dataset is refined to 6,361,455 data samples.

\begin{figure}[h]
\centering
\vspace{0pt}
\includegraphics[trim=0pt 0pt 0pt 0pt, clip, width=4.3in]{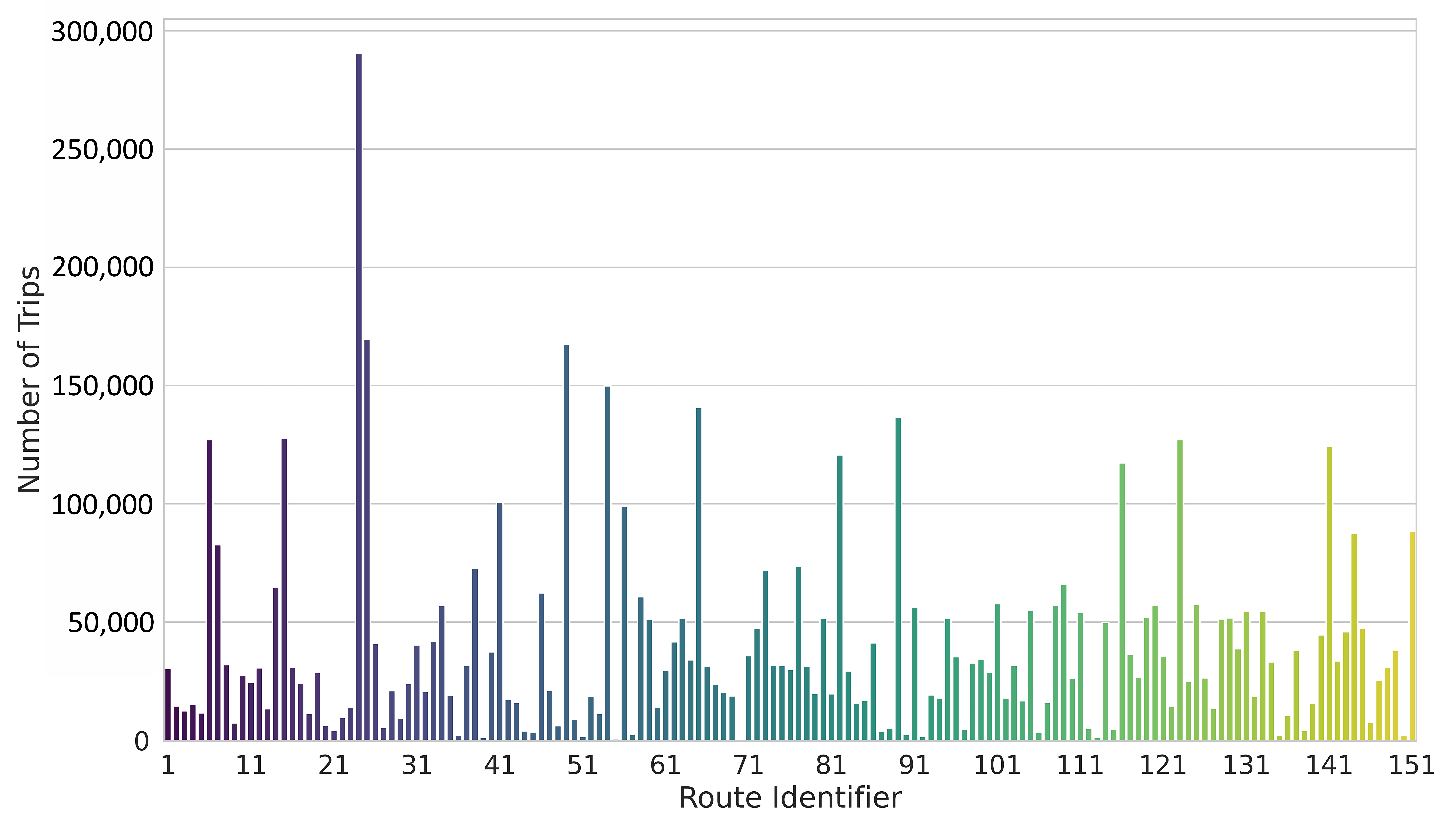}
\caption{The distribution 
 of trips acros\textls[-25]{s the routes over the three months ranges from a maximum of 290,545~trips to a minimum of 568 trips}.}
\label{fig_distribution}
\end{figure}

As noted earlier, the dataset includes 151 routes. Figure \ref{fig_distribution} illustrates how trips are distributed across these routes, ranging from a peak of 290,545 trips to a minimum of 568 trips over the three-month period.

Next, the statistical analysis examines the variations between scheduled and actual bus departure times, known as departure time deviation. To quantify this, a variable named "current stop departure time deviation" is introduced, which measures the time difference between actual and scheduled times at each stop. Over a three-month period, this feature ranges from $-$16139 to 35558. This indicates that the maximum early departure is 16,139 seconds ahead of schedule, while the longest delay recorded is 35,558 seconds. The next step involves identifying outliers, which are systematically detected using statistical methods: 
\begin{equation} \label{eqn1}
	 \text{High Threshold} = M + k \sigma 
	\end{equation}
 \begin{equation} \label{eqn2}
    \text{Low Threshold}= M - k \sigma
	\end{equation}
where $M$ is the mean, and $\sigma$ is the standard deviation, with values of 261.84 and 309.996, respectively. For this study, $k$ was set to 2, aligning with the range within which approximately 95\% of normally distributed data fall. Values beyond these thresholds were flagged as outliers due to their extreme deviation from the mean.
The calculated thresholds, based on normal distribution principles, were 881.832 for the high threshold and $-$358.152 for the low threshold, providing clear boundaries for outlier detection within the dataset.
After removing outliers, as shown in Table \ref{tab:dataset_characteristics}, a total of 406,969 data points were identified and subsequently excluded. Following this cleanup, the refined dataset now shows a mean departure time deviation of 211.866 across all bus lines included in the analysis.

\begin{figure}[h]
\centering
\vspace{0pt}
\includegraphics[trim=0pt 0pt 0pt 0pt, clip, width=4.3in]{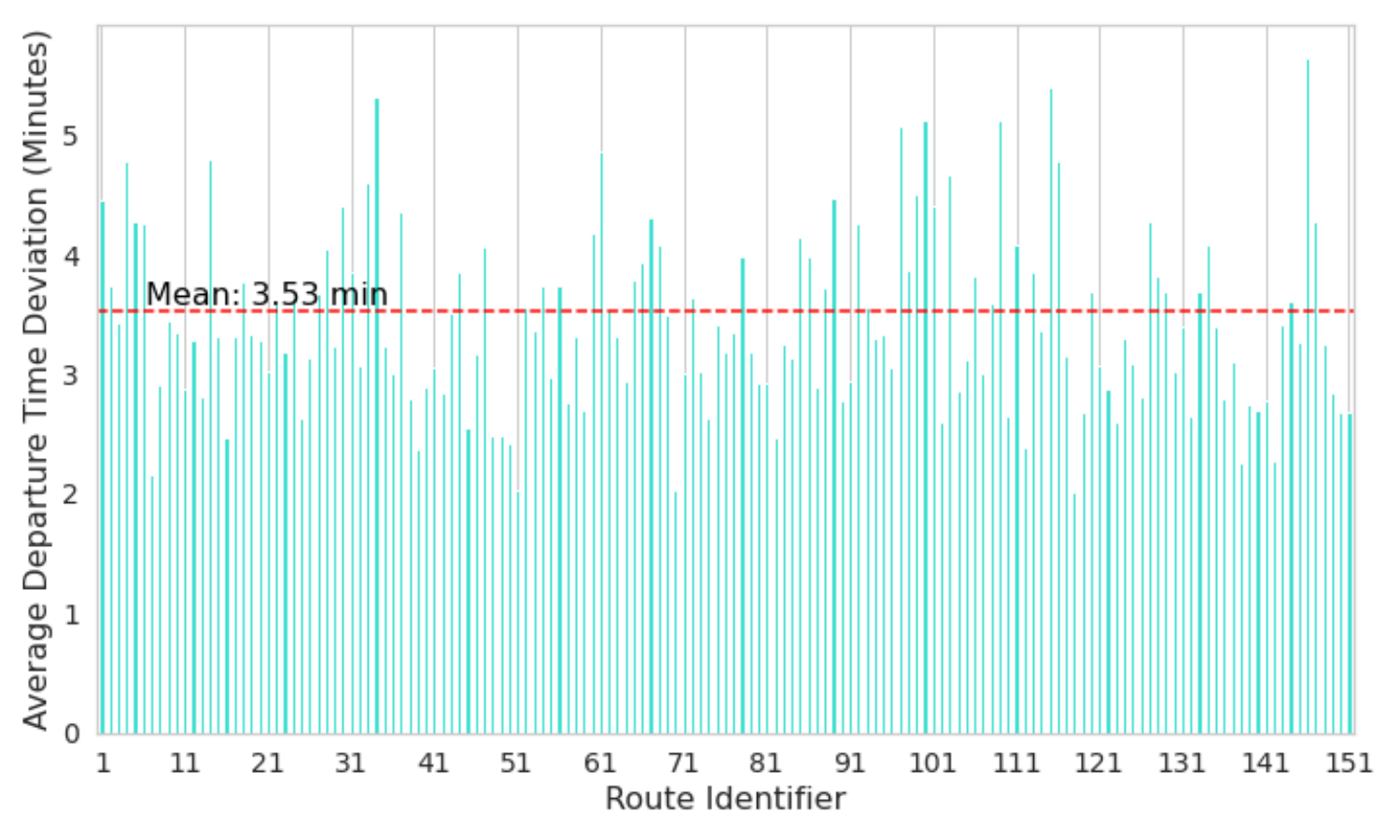}
\caption{Average departure time deviation across routes in  preprocessed MBTA bus departure times 2023 dataset.}
\label{fig_time_deviation_dataset}
\end{figure}

Our research revealed that the average departure time deviation across all routes, defined as the average difference between the scheduled departure times and the actual departure times,  in the preprocessed dataset is 3 minutes and 31 seconds, as illustrated in Figure \ref{fig_time_deviation_dataset}.
Out of the total number of trips, 4,585,512 are delayed, while the remainder are ahead of schedule. This results in delayed trips accounting for almost 77\%, with ahead-of-time trips making up the remaining 23\%.

\subsection{Feature Extraction and Data Preparation for Model Implementation}
\begin{figure}[h]
\centering
\vspace{0pt}
\includegraphics[trim=20pt 320pt 20pt 200pt, clip, width=4in]{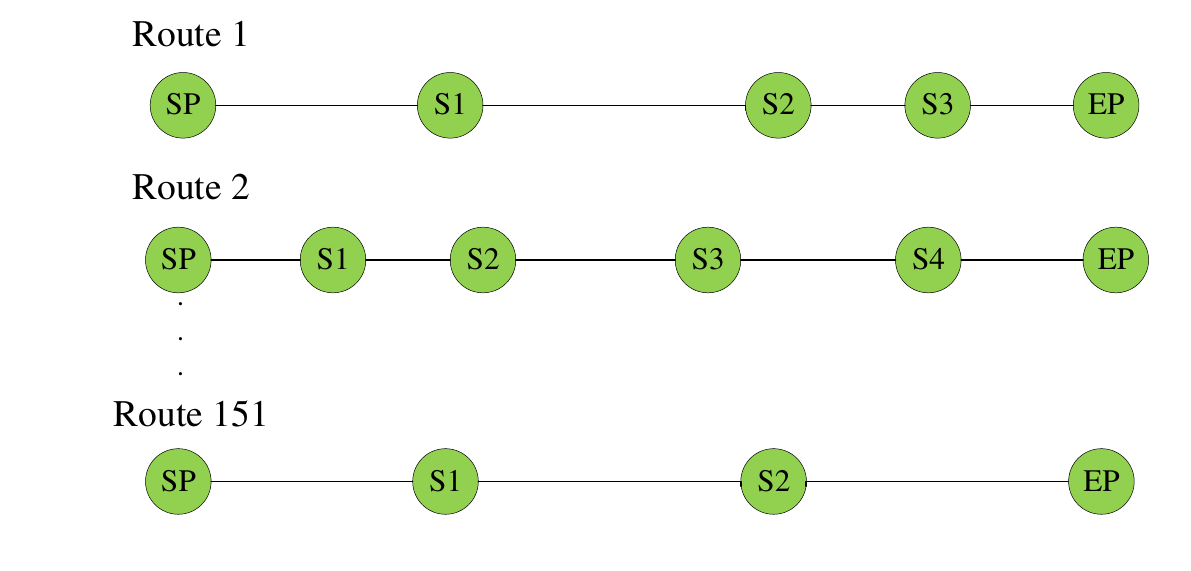}
\caption{Segmenting each trip into sequential segments of bus stops from the start point (SP) to the endpoint (EP), with each trip containing between 2 and 14 stops.}
\label{fig_segmentation}
\end{figure}

After cleaning the dataset, relevant features are generated, which will be detailed in this section.
In general, each bus route is segmented into multiple bus stops. This segmentation is carried out by utilizing the "Half-Trip ID" field to distinguish individual trips. Upon grouping the trips by "Half-Trip ID", we observe that the dataset includes trips ranging in length from a minimum of 1 stop to a maximum of 14 stops in length. To focus on predicting departure time based on the previous stop, trips consisting of only one stop are excluded from the dataset, while retaining all other data points. Consequently, the dataset now comprises trips with 2 to 14 stops along their routes. 
Furthermore, to prepare the dataset for algorithm development, we use the "Stop ID" and "Timepoint Order" fields to establish the sequential order of each stop along the trip route.  We then segment each trip into smaller segments between every pair of bus stops, from the start point (SP) to the endpoint (EP), as illustrated in Figure \ref{fig_segmentation}.

After preparing the dataset for model implementation by grouping trips and establishing the sequential order of stops, the next step involves feature engineering. The input features for the model are categorized into four groups, time-related features, distance-related features, weather-related features, and traffic-related features, with the output feature being the next stop departure time deviation, as illustrated in Figure \ref{FlowChart} and explained below.

\label{sec:data}  
\begin{itemize}
\item Time-related features:

As previously mentioned, the bus departure records from this dataset were collected over three months, resulting in a wide range of time variations. To minimize potential noise, time is categorized into two distinct features rather than being directly incorporated. Firstly, the variable "day type" is introduced into the input features, distinguishing between "weekend" and "workday" operations based on the day of bus operation.

Additionally, another time-related variable, "rush hour status", identifies whether each record corresponds to bus operations during rush hour. Rush hour in Boston is defined as the period from 7 AM to 9 AM and from 4 PM to 6 PM \cite{Stopslocation}. 
We also include another input variable, "current stop lateness status", which indicates whether the bus is behind or ahead of schedule. This variable is derived from the status of the previous stop and is determined by whether the deviation from the scheduled departure time at the current stop is positive or negative. Additionally, the model incorporates the current stop departure time deviation as an input, calculated as the difference between the actual departure time and the scheduled departure time, allowing it to assess the degree of delay or advancement in the bus schedule.

\item Distance-related features:

Our model incorporates two distance-related features. The first feature is the "stop-to-stop distance", which quantifies the distance in meters between consecutive bus stops. This distance is calculated using the X and Y coordinates of both the current and next stop. The second variable, termed "far status", is a binary indicator based on a threshold derived from the mean distance between bus stops in Boston, reported as 1488 meters according to \cite{Stopslocation}. In this dataset, the average distance between stops is 1233.26 meters. This binary variable reflects whether the distance between two consecutive bus stops exceeds or falls below this threshold.

\item Weather-related features:

To handle weather-related features, categorized as conditions including cloudy, rainy, clear day, snowy, and windy, we employ one-hot encoding. This method converts each weather condition into binary variables, thereby adding 5 additional variables to the input features of the model.

\item Traffic-related features:

One-hot encoded inbound and outbound directions, as well as Route IDs, are the features categorized in this group. With 151 routes in the dataset, this approach introduces 151 input features for routes and 2 input features for directions. Furthermore, the model incorporates the X and Y coordinates of the current and next bus stop, enabling the analysis of traffic patterns across diverse geographical areas.

\item Output variable:

The final target variable in our model is the next stop departure time, as illustrated in Figure \ref{BusModels}. Since the scheduled departure time for each stop is available, the algorithm is designed to predict the departure time deviation of the next stop. Assuming that bus $n$ is currently between stop $i$ and $i+1$, the predicted departure time deviation for the next stop allows us to compute the actual departure time using the following formula:

\begin{equation} \label{eqn}
	{t_{pred}(n, i+1)} = {t_{sched}(n, i+1)} + {d_{dep}(n, i+1)}
	\end{equation}  
where ${t_{pred}(n, i+1)}$ represents the predicted departure time of bus $n$ at the next bus station ($i+1$), ${t_{sched}(n, i+1)}$ is the scheduled departure time of this bus at the next station, and  ${d_{dep}(n, i+1)}$ is the predicted departure time deviation for the next stop.

\begin{figure}[h]
\centering
\vspace{0pt}
\includegraphics[width=\linewidth]{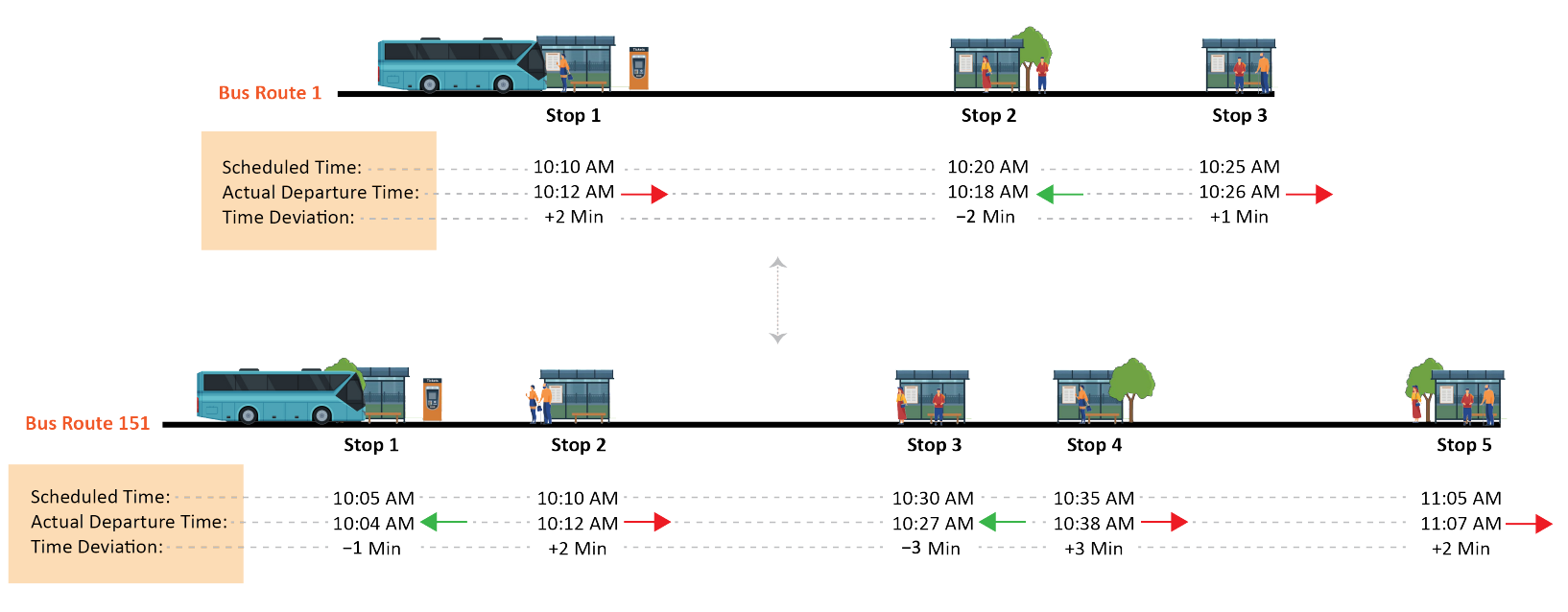}
\caption{The 
 image 
 illustrates the application of a bus departure time prediction model using an FCNN for two bus routes. It shows the scheduled and actual departure times, highlighting deviations with colored arrows. Red right-facing arrows indicate late departures, and green left-facing arrows indicate early departures.}
\label{BusModels}
\end{figure}

\end{itemize}

\section{Experimental Results and Discussion} \label{sec5}
\subsection{Network Architectures}
\label{sec:intro}  
Artificial neural networks (ANNs) are used in bus scheduling to predict trip times and arrival/departure times due to their effectiveness in capturing non-linear relationships in complex problems \cite{rashvand2023real}. 

\begin{figure}[h]
\centering
\vspace{0pt}
\includegraphics[width=4.6in]{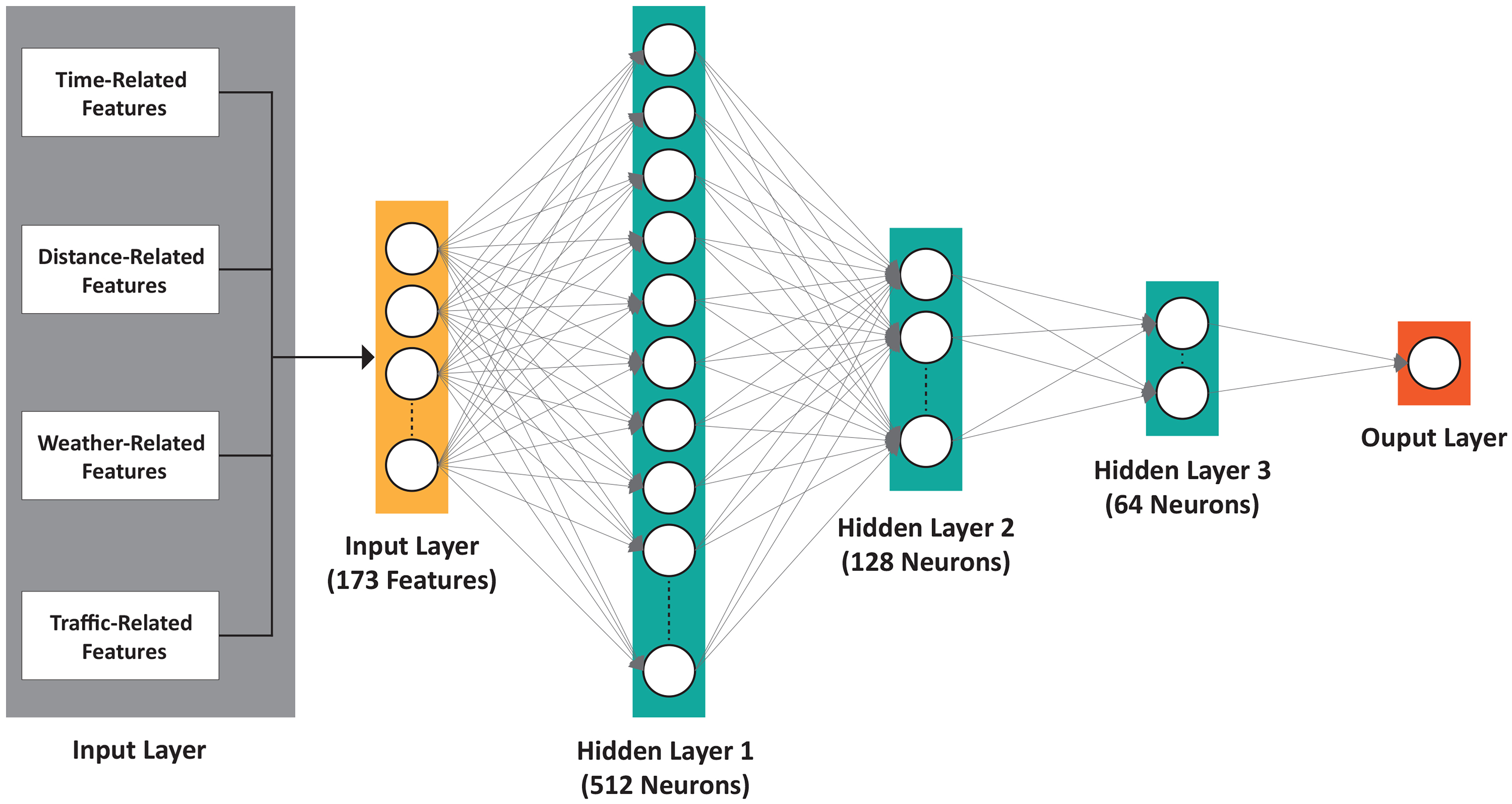}
\caption{The model processes 173 input features, including time, distance, weather, and traffic data, through its hidden layers. The illustrated structure represents the model with optimal performance in terms of accuracy and computational complexity, with three hidden layers comprising 512, 128, and 64 neurons, respectively. The final output layer provides departure time deviation predictions.}
\label{Neuralnet}
\end{figure}

In this study, we utilize FCNNs to predict the next stop bus departure time deviation. FCNNs excel in handling high-dimensional feature spaces through the use of hidden layers and non-linear activation functions. As depicted in Figure \ref{Neuralnet}, the model incorporates various input features, incorporating information from the current stop and other relevant features, including those derived from the feature extraction step. Specifically, each data point includes 173 input features and the output layer comprises a single neuron responsible for predicting the departure time deviation at the next stop. Throughout our experiments, the configuration of the input and output layers remained consistent.

To ensure effective handling of input features with diverse ranges, preprocessing involves scaling the data. For instance, features like rush hour status are binary, while distance can vary across hundreds of meters. Without scaling, the model prioritizes features with large numerical ranges. Therefore, we apply Min-Max scaling to normalize all input features to a range between 0 and 1. It is important to note that this scaling was applied exclusively to the input variables.
To optimize our model architecture, we conducted extensive experiments by varying parameters such as the number of hidden layers, the number of neurons per layer, and activation functions, as detailed in the model performance section. Through these experiments, we identified the most effective configuration for achieving accurate predictions.

\subsection{Metrics}
In this study, we use the Root Mean Square Error (RMSE) to measure the difference between predicted and actual departure time deviation in seconds. The RMSE is commonly used in the analysis of bus arrival/departure times. Furthermore, the RMSE is expressed in the same unit as the predicted values (seconds), facilitating a clear interpretation of error in terms of time \cite{rashvand2023real}. 
The RMSE is calculated using the following equation, where $d_{act}(n, i+1)$ represents the actual departure time deviation for the next stop of bus $n$, $d_{pred}(n, i+1)$ denotes the predicted departure time deviation of the next stop, and $m$ is the sample size for prediction.

\begin{equation} \label{eqn_RMSE}
	RMSE=\sqrt \frac {\Sigma^{m}_{i=1}  (d_{act}(n, i+1) - d_{pred}(n, i+1))^2} {m}
	\end{equation}

 \subsection{Model Performance}
\label{sec:result} 

For evaluating the performance of our proposed model, we utilized the preprocessed MBTA 2023 dataset. This dataset was randomly divided, allocating 70\% for training, 20\% for validation, and 10\% for testing.

Our experimental setup was based on the PyTorch framework, and we employed the MSE loss function. Furthermore, all experiments were performed on a GPU server equipped with 4 Tesla V100 GPUs with 32 GB memory. The Adam optimizer was used for all experiments. Each model underwent training for 10 epochs, with a learning rate of 0.01.

In our experimental approach, we extensively explored the FCNN architecture by systematically varying the number of hidden layers from one to five and examined their impact on model performance. We evaluated how different configurations influenced model performance and model complexity.
We adjusted the number of neurons in descending order, with the first layer having the highest number of neurons and the final layer having the fewest. 
This adjustment was aimed at finding an optimal balance between model complexity, generalization, and accuracy, ensuring that the FCNN-based model effectively captures complex data patterns while avoiding overfitting and unnecessary parameters.

Table \ref{tab:ablation} provides a detailed overview of the model's architecture and specifications from the ablation study, including the number of hidden layers and the number of neurons in each layer from input to output. In this table, "H1" refers to the first hidden layer, with the number of neurons in that layer listed immediately afterward.
\begin{table}[htpb]  
\centering
\caption{A comparative analysis of various architectures based on Test RMSE, number of parameters, and computational complexity using the preprocessed MBTA Bus Departure Times 2023 dataset. The optimal architecture is highlighted in bold.}
\label{tab:ablation}
\resizebox{\textwidth}{!}{%
\begin{tabular}{l|c|c|c}
\toprule[\heavyrulewidth] \midrule
\textbf{Architecture Details}  
& \textbf{Test RMSE}  
& \textbf{Number of Parameters}  
& \textbf{Computational Complexity (FLOPs)} \\ 
\midrule
One hidden layer (H1-256N) & 87.2219 & 44.801 K & 44.544 M \\
Two hidden layers (H1-256N, H2-32N) & 80.9800 & 52.801 K & 52.512 M \\
Two hidden layers (H1-256N, H2-64N) & 79.5718 & 61.057 K & 60.736 M \\
\textbf{Three hidden layers (H1-512N, H2-128N, H3-64N)} & \textbf{77.8312} & \textbf{163.073 K} & \textbf{162.368 M} \\
Four hidden layers (H1-1024N, H2-512N, H3-64N, H4-32N) & 76.8411 & 737.921 K & 736.288 M \\
Five hidden layers (H1-512N, H2-256N, H3-128N, H4-64N, H5-32N) & 76.8323 & 263.681 K & 262.688 M \\
Five hidden layers (H1-1024N, H2-512N, H3-128N, H4-64N, H5-32N) & 76.9803 & 779.009 K & 777.248 M \\ 
\midrule \bottomrule[\heavyrulewidth]
\end{tabular}%
}
\vspace{-20pt}
\end{table}

Computational complexity and model size are crucial considerations, particularly for IoT devices with limited storage and computational resources. Computational complexity can be measured through various metrics, including FLOPs and Multiply--Accumulate Operations (MACs). In this study, we evaluate the computational complexity using FLOPs and assess the model size based on the number of parameters. FLOPs represent the total number of floating-point operations required by the model, offering a clear indication of its computational load.
Table \ref{tab:ablation} demonstrates that increasing the complexity of the model architecture by adding more hidden layers and increasing the number of neurons per layer leads to a rise in both the total number of parameters and FLOPs, as expected. However, this increased complexity does not consistently translate to improved performance in terms of the RMSE. For instance, an FCNN with a single hidden layer yields an RMSE of 87.2219~s. In contrast, the RMSE decreases to approximately 77.8312 seconds when the model is expanded to three hidden layers, reflecting an 11\% reduction in prediction error, though with a fourfold increase in the number of parameters and FLOPs. Beyond this point, further increasing model complexity results in diminishing returns, with RMSE improvements leveling off and offering minimal enhancement in overall performance. As illustrated in the table, the RMSE decreases by less than 1\% when expanding the model from three to five hidden layers. Given the minimal performance gains and the significant increase in computational cost and complexity associated with additional layers, we chose not to extend the model beyond five hidden layers in our design. Therefore, we selected the model with one input layer, three hidden layers, and one output layer as the optimal configuration for the remainder of our experiments. The detailed structure of this model is provided in Table \ref{tab:ablation}.

{We also employed the regression model to compare its performance against the
FCNN-based model introduced earlier. The regression model, which comprises 174 parameters, lacks non-linear activation functions such as ReLU, resulting in an RMSE of 161.4434~s, which is 2.07 times higher than the RMSE of the optimal model identified in our ablation study. A key limitation of traditional models, like regression and support vector regression (SVR), is their inability to capture complex, non-linear relationships within the data. In contrast, the FCNN effectively models these complexities through its multiple layers and non-linear activation functions, as highlighted in the field of arrival/departure time prediction for transportation systems \cite{yin2017prediction,rashvand2023real}. As a result, the optimal model from our ablation study demonstrates significantly greater accuracy and robustness in this context.}

{Moreover, we measured both the MAE and MAPE for our optimal model. The MAE, which calculates the average absolute difference between the actual and predicted values, is expressed in the same unit as the target variable. For our test set, the MAE using the optimal model is 55.63 seconds.
To calculate the MAPE, we excluded about 620 data points where the departure time deviation was zero, as including these would lead to infinite percentages. We then calculated the MAPE for the remaining test set, yielding a value of 63.65\%. However, this metric requires further clarification in the context of our problem. For instance, let us consider a data point where the actual departure time deviation is 95 seconds, and the model predicts a deviation of 136 seconds. Although the absolute error is 41 seconds, the MAPE is 43.16\%. 
Despite the fact that a 95-second departure time deviation is relatively minor and a prediction of 136 seconds is quite close, the percentage error seems large due to the relative nature of the MAPE. 
Moreover, let us consider these two data points in the test set. In one, the actual value is 726 seconds, with a prediction of 706 seconds, resulting in an MAPE of 2.75\%. This indicates a small error, reflecting good model performance. In contrast, for an actual value of 37 seconds and a prediction of 17.32 seconds, the MAPE rises to 53\%. Here, the absolute error is still 20 seconds, which is acceptable, but the MAPE inflates this error due to the small actual value and suggests a significant error in the model's prediction when viewed in percentage terms.
These scenarios illustrate how the MAPE may not be a suitable metric for evaluating deviation prediction and how it can misrepresent model performance in this context with the dataset that originally has a departure time deviation of almost 4 minutes, as shown in Figure \ref{fig_time_deviation_dataset}. The MAPE's sensitivity to the small actual value means that even a modest absolute error can result in a high percentage error. Thus, it is essential to use additional metrics to gain a complete picture, and relying solely on the MAPE in this field can obscure important insights.}  

\begin{figure}[h]
\centering
\vspace{0pt}
\includegraphics[trim=0pt 0pt 0pt 0pt, clip, width=4.2in]{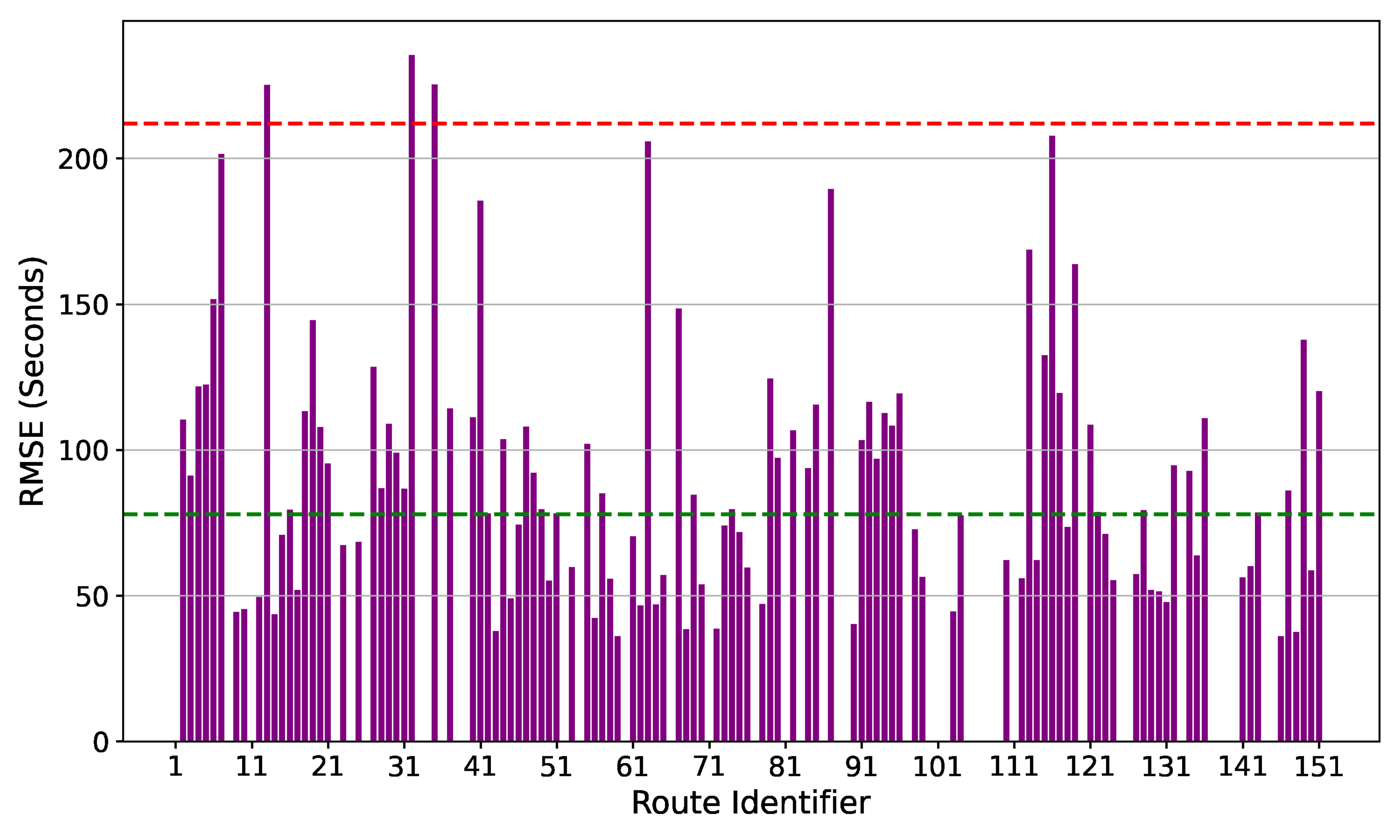}
\caption{The performance of the proposed model across all bus routes in the preprocessed MBTA bus departure times 2023 test set. The average RMSE for all routes in the test set is 77.8312 s. Among the prediction errors, bus route 32 has the greatest RMSE, with a value of 240.61 s.}
\label{fig_results}
\end{figure}

Furthermore, we extended our analysis by measuring the RMSE for each bus line based on the test set using the optimal model. Figure \ref{fig_results} illustrates the RMSE for each bus line, with the dataset's average departure time deviation indicated by a red line and the average RMSE across all bus lines represented by a green line. The average departure time deviation as discussed before is 211.866 s, while the average RMSE for the test set is 77.8312 s.
Among the bus routes in the test set, the highest prediction error is observed for bus line 32, with an error of 240.61 s.

The higher RMSE values observed for some bus lines compared to others may result from a lack of relevant features specific to those routes. To improve the accuracy of departure time predictions, incorporating additional relevant features is necessary. For example, factors like passenger demand, which influence how long buses stay idle at stops (known as dwell times), are not included in the dataset. Integrating passenger counting systems on buses and incorporating features related to passenger demand at each stop could enhance the model's predictive accuracy. Thus, investigating the impact of passenger demand on prediction accuracy could be a promising area for future research. We also measured the inference time per test data point, which is 28.7 microseconds. This indicates that predictions for bus departure times are generated in 28.7 microseconds when requested by a passenger. However, it is important to clarify that this inference time refers to the model's processing time based on its inputs and does not encompass communication network latency or other delays. Additionally, this time relates to individual access request. Discussions of overall response time, particularly as the number of passengers requesting bus departure times in a major metropolitan area increases, fall outside the scope of this paper, which focuses exclusively on the data engineering and algorithmic aspects. Nonetheless, it is worth noting that increased passenger demand will significantly impact the total response time.

\section{Conclusions} \label{sec6}
\label{sec:conclusion} 
{In this study, we developed an AI-driven prediction model designed for an intelligent bus transit system to enhance the experience of both passengers and service providers. Our main goal was to improve the overall efficiency and reduce waiting times for passengers. 
We utilized three key datasets, the Boston MBTA Bus Departure Times 2023 dataset, the Boston weather dataset, and the MBTA Bus Routes and Stops dataset, to extract and engineer relevant features. This study highlights the adaptability of the FCNN architecture in capturing temporal patterns through effective data handling and analysis of previous bus stop data. We conducted a thorough examination of FCNN performance across various configurations, varying in the number of hidden layers and neurons. In our case study analyzing bus data from Boston, we found an average deviation of nearly 4 minutes from scheduled times. However, our model, evaluated across 151 bus routes, achieved a significant improvement, predicting departure time deviation with an accuracy of within 80 seconds. Our findings underline the importance of balancing model size and performance, crucial for deployment in resource-constrained environments like IoT devices. The comparative analysis demonstrates the model's potential in achieving both high accuracy and parameter efficiency.
Moving forward, there are several numerous paths for further exploration in this field. First, investigating the incorporation of additional factors such as passenger flow could offer a more comprehensive insight into the variables impacting bus departure times. Additionally, since the dataset covers an entire year, future research could explore more complex architectures and include seasonal features in the feature engineering process. Furthermore, investigating advanced architectures, particularly transformers and autoencoders, presents a valuable approach to improve our model's performance. Transformers use self-attention to effectively capture long-range dependencies within the data, while autoencoders excel at learning compact representations and minimizing noise. Using these models can enhance the accuracy and efficiency of predictions in transportation systems, which is a key focus for our future research.}

{\large \textbf{Acknowledgment}}

This study is supported by the UNC Charlotte's College of Engineering seed grant.

\bibliographystyle{unsrtnat}
\bibliography{references}  






\end{document}